\documentclass{article}
\usepackage{ijcai20}

\usepackage{times}
\usepackage{soul}
\usepackage{url}
\usepackage[hidelinks]{hyperref}
\usepackage[utf8]{inputenc}
\usepackage[small]{caption}
\usepackage{graphicx}
\usepackage{amsmath}
\usepackage{amsthm}
\usepackage{booktabs}
\usepackage{algorithm}
\usepackage{algorithmic}
\usepackage{times}
\usepackage{graphicx}
\usepackage{latexsym}
\usepackage[utf8]{inputenc}
\usepackage[english]{babel}
\graphicspath{{images/}{../images/}}
\usepackage{import}
\usepackage{mathptmx}
\usepackage[dvipsnames]{xcolor}
\usepackage{lipsum} 
\usepackage{graphicx}
\usepackage{amsmath}
\usepackage[toc,page]{appendix}
\usepackage{tabularx}
\usepackage{tabulary}
\usepackage{tabularht}
\usepackage{hyperref}
\usepackage{subcaption}
\usepackage{listings}
\usepackage{adjustbox}
\usepackage{multirow}
\usepackage{array}
\title{Towards Ranking Schemas by Focus}

\author{
Mattia Fumagalli$^1$\footnote{Contact Author}\and
Daqian Shi$^1$\and
Fausto Giunchiglia$^1$
\affiliations
$^1$DISI - University of Trento\\
\emails
\{mattia.fumagalli, daqian.shi, fausto.giunchiglia\}@unitn.it
}

\begin{document}

\maketitle

\begin{abstract}
The main goal of this paper is to evaluate \textit{knowledge base schemas}, modeled as a set of entity types, each such type being associated with a set of properties, according to their \textit{focus}. We intuitively model the notion of focus as \textit{``the state or quality of being relevant in storing and retrieving information''}. This definition of focus is adapted from the notion of \textit{``categorization purpose''}, as first defined in cognitive psychology, thus giving us a high level of understandability on the side of users. In turn, this notion is formalized based on a set of knowledge metrics that, for any given focus, rank knowledge base schemas according to their quality. We apply the proposed methodology on more than 200 state-of-the-art knowledge base schemas. The experimental results show the utility of our approach.\footnote{Code and data will be released upon acceptance.} 
\end{abstract}

\section{Introduction}
Following contemporary descriptions by psychologists, the purpose of what we call categorization can be reduced to a ``\textit{...a means of simplifying the environment, of reducing the load on memory, and of helping us to store and retrieve information efficiently}'' \cite{klapper2017four,rosch1999principles,harnad2017cognize}. 
Loosely speaking, categorizing consists of putting things (events, items, objects, ideas, people) into categories (e.g., classes or types) based on their similarities, or common features. Without categorization we would be overwhelmed by the huge amount of diverse information coming from the external environment and our mental life would be chaotic \cite{millikan2000clear,giunchiglia2016concepts}. 

In AI the purpose of categorization is usually implemented by well-defined and effective information objects, namely knowledge bases schemas (KBSs); prominent examples include \textit{knowledge graphs (KGs) schema layers} \cite{qiao2016knowledge} and \textit{ontologies} \cite{guarino2009ontology}. KBSs offer many pivotal benefits \cite{paulheim2017knowledge}, such as: i) human understandability; ii) a fixed and discrete view over a stream of multiple and diverse data; iii) a tree or a grid structure, so that each information can be located by answering a determinate set of questions in order; and iv) an encoding in a formal language, which is a fragment of the first first-order predicate calculus. 
These benefits allow to represent high-performance solutions to large scale categorization problems, namely problems of efficient information storage and retrieval.

KBSs are the backbone of many semantic applications and play a central role in improving the efficiency of many ``categorization systems'' (like digital libraries or online stores). Unfortunately, their construction involves a very huge effort in terms of time and domain specific knowledge (see for instance well-known problems as ``knowledge acquisition bottleneck'' \cite{shadbolt2015knowledge}). So far, in order to minimise the effort in building KBSs, a huge number of search engines, catalogs and metrics have been produced, to facilitate their reuse \cite{mcdaniel2019evaluating}. The facilities provided by these solutions are very valuable especially in the context of their \textit{structural evaluation}. However, there is still a lot of work to be done for their \textit{functional, or purpose-driven, evaluation}, in particular to what concerns their ranking based on the \textit{relevance\footnote{``\textit{Something (A) is relevant to a task (T) if it increases the likelihood of accomplishing the goal (G), which is implied by T''} \cite{hjorland2002work}.} of their concepts, and their categorization purpose} \cite{butt2016dwrank}. The main motivation behind this situation is that most of the purpose-driven features of KBSs are extremely difficult to \textit{quantify} and they are very context dependent \cite{gangemi2005theoretical}. Just consider, for instance, the essential role of the qualifier ``core'' for describing ontologies and its multiple possible interpretations \cite{falbo2016ontology}). As the number of available KBSs increases, this issue is bound to get worse. There is an increasing need for more and more KBSs to be constructed and made available, and, as this situation occurs, their purpose-driven evaluation for facilitating their reuse, becomes an even greater problem. 

The main goal of this paper is to evaluate KBSs, modeled as a set of \textit{entity types}, each such type being associated with a set of \textit{properties}, according to their  \textit{focus}, where we intuitively model the notion of focus as ``\textit{the state or quality of being relevant in storing and retrieving information}''. This definition of focus is adapted from the notion of \textit{categorization purpose}, as first defined in cognitive psychology, thus giving us a high level of understandability on the side of users. 

In order to support an accurate level of KBSs reuse, we then propose a solution to rank KBSs that is articulated into three main contributions: 

\begin{enumerate}
\item a cognitive psychology grounded account of the notion of focus (Section \ref{sec1});
\item a set of metrics that apply to KBSs, their entity types, and their properties, which can be used to rank KBSs according to their focus (Section \ref{sec2});
\item an analysis of the application of the metrics over a huge amount of state-of-the-art (SoA) data sets (Section \ref{sec3}). 
\end{enumerate}

The evaluation (Section \ref{sec4}) confirms the validity of the approach. Section \ref{sec5} discusses the related work, Section \ref{sec6} the conclusions.

\section{Defining focus} \label{sec1}
Imagine that by saying ``the green book on my desk in my office'' someone wants someone else to bring her that book. This will happen only if the two subjects share \textit{a way of describing objects} into those that are offices and those that are not, those that are books and those that are non-books, green things and non-red things, desks and non-desks. These ``object descriptions'' are what is meant to convey for retrieving the intended objects. The point is \textit{to draw sharp lines around the group of objects to be described}. That is \textit{the categorization purpose} of an object description. These object descriptions, also called types, categories or classes, are the basis of the organization of our mental life. Meaning and communication heavily depend on this categorization \cite{millikan2000clear,millikan2017beyond,fumagalli2019towards}. 

Following the contemporary descriptions by psychologists, and, in particular, the seminal work by Eleanor Rosch \cite{rosch1999principles}, the categorization purpose of objects descriptions or categories, can be explained according to two main dimensions, namely: \textit{i)} \textit{the maximization of the number of features that describe the members of the given category} and \textit{ii)} \textit{the minimization of the number of features shared with other categories.}

To evaluate these dimensions Eleanor Rosch introduces the central notion of \textit{cue validity} \cite{rosch1975family}. This notion was defined as ``\textit{the conditional probability } $p(c_{j} | f_{j})$ \textit{that an object falls in a category} $c_{j}$ \textit{given a feature}, \textit{or cue}, $f_{j}$'', and then used to define the set of basic level categories, namely those categories which maximize the number of characteristics (i.e., features or attributes like ``having a tail'' and ``being colored'') shared by their members and minimize the number of characteristics shared with the members of their sibling categories. The intuition is that \textit{basic level categories} have higher cue validity and, because of this, they are \textit{more relevant in categorization}.

Rosch's definitions were designed for experiments where humans were asked to score objects as members of certain given categories. We adapt Rosch's original methodology to the context of knowledge base schema design. In our setting, each available KBS (see, for instance, \textit{schema.org}\footnote{\url{http://schema.org/}} or \textit{DBpedia}\footnote{\url{https://wiki.dbpedia.org/}}) plays the role of a categorization, which is modeled as a set of \textit{entity types} associated to a set of \textit{properties}, whose main function is to \textit{draw sharp lines around the types of entities it contains, so that each member in its domain falls determinately either in or out of each entity type} \cite{giunchiglia2020entity,2019jowo}. The knowledge engineers play a similar role of the persons involved in Rosch's experiment. Each knowledge base schema provides a rich set of categorization examples. Each entity type plays the role of a category and all entity type properties play the role of features. The categorization purpose of the KBS is then mapped into the notion of \textit{focus}. 
We intuitively model the notion of focus as ``\textit{the state or quality of being relevant in storing and retrieving information}'' and we quantify the degree of this relevance by adapting the Rosch's notion of cue validity as follows:

\begin{itemize}
\item we take each property has having the same ``cue validity'' (which we assume to be normalized to one);
\item for each KBS we equally divide the property ``cue validity'' across the entity types the properties are associated to;
\item by checking the wide-spreading of ``cue validity'' we quantify the relevance of the KBS and entity types in storing and retrieving information.   
\end{itemize}

The ``focus'' can be then calculated in relation to this analysis and, in turn, it can be functionally articulated in:
\begin{itemize}
\item \textit{the entity types focus}, namely, what allows to identify the entity types that are maximally informative categories, which have a \textit{higher categorization relevance}, or, more precisely, which maximise the amount of properties and minimize the number of properties shared with other categories. These entity types are, in some extent, related to what expert users consider as ``core entity types'' or central entity types for a given domain; 
\item \textit{the KBSs focus}, namely, what allows to identify the KBSs that maximise the number of maximally informative (focused) entity types. These KBSs being described, in some extent, as \textit{``clean''} or \textit{`not-noisy''} and are related to what expert users classify as well-designed KBSs \cite{paulheim2017knowledge}.
\end{itemize}

\section{Focus metrics}
\label{sec2}
\begin{figure*}[ht]
  \centering
  \includegraphics[width=\linewidth]{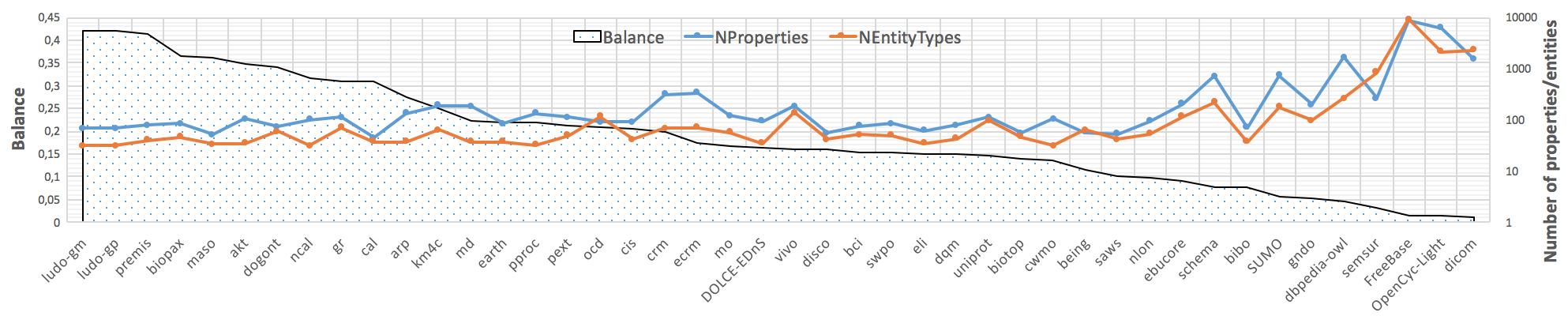}
  \vspace{-1.5em}
  \caption{\textit{KBSs selected for the analysis}}
  \label{01}
  \vspace{-1.5em}
\end{figure*}

We assume that a KBS can be formalized as: $K = \left \langle E_{K},P_{K},I_{K} \right \rangle$, with $E_{K} = \left \{ e_{1}, ..., e_{n} \right \}$ being the set of \textit{entity types} of \textit{K}, $P_{K} = \left \{ p_{1}, ..., p_{n} \right \}$ being the set of \textit{properties} of \textit{K}, and $I_{K}$ being a binary relation $I_{K} \subseteq E_{K} \times P_{K}$ that expresses which entity types \textit{are associated} to which properties. We say that an entity type is \textit{associated to} a property when the latter is used to describe the former, and that a property is \textit{associated to} an entity type with the dual meaning. We also talk of an entity \textit{e} being in the domain of a property \textit{p}, in formulas {e} $\in$ \textit{dom(p)},  when \textit{e} is associated with \textit{p}. Thus, for instance, the entity \textit{Person} can be in the domain of properties such as \textit{address} or \textit{name}, while the property address may be associated with entities such as \textit{Person}, or \textit{Building}. 
Notice how, the proposed formalization of entities and properties is different from, e.g., the OWL\footnote{https://www.w3.org/2001/sw/WebOnt/} representational language. The key difference can be clarified by considering our formalism as very similar to what is proposed by the \textit{Formal Concept Analysis} (FCA) methods \cite{ganter2012formal}. Our commitment to this model is motivated not only on foundational considerations, but also on pragmatical grounds. In fact, once properties and entities are \textit{encoded} as described above, data can be analyzed and processed with very few limitations.\footnote{See \cite{goyal2018graph} for an overview of the multiple available approaches and applications}

Given the above formalization we define a main set of metrics to evaluate KBSs according to their focus. 

Firstly, we define the \textit{cue validity of a property p w.r.t to an entity e}, also called $cue_{p}-validity$, as: 

\vspace{-0.2cm}
\begin{equation}
Cue_{p}(p, e) = \frac{PoE(p,e)}{|dom(p)|} = c \in [0, 1]
\vspace{-0.05cm}
\end{equation}

with $\mid \textit{X} \mid$ being the cardinality of the set \textit{X} and \textit{PoE(p, e)} being defined as:

\vspace{-0.2cm}
\begin{equation}
PoE(p,e) = \left\{\begin{matrix}1, if e \in dom(p)
\\ 
0, otherwise
\end{matrix}\right.
\vspace{-0.05cm}
\end{equation}

$Cue_{p}(p, e)$ returns 0 if \textit{p} is not associated with \textit{e} and \textit{1/n}, where \textit{n} is the number of entities in the domain of \textit{p}, otherwise. In particular, if \textit{p} is associated to only one entity type its $cue_{p}-validity$ is maximum and equal to one.

Given the notion of $cue_{p}-validity$ we define the notion of cue validity of an entity type, also called $cue_{e}-validity$, as the sum of the cue validities of the properties associated with the entity, namely:

\vspace{-0.2cm}
\begin{equation}
Cue_{e}(e) = \sum_{i=1}^{|prop(e)|} Cue_{p}(p_{i}, e) = c \in [0, |prop(e)|]
\vspace{-0.05cm}
\end{equation}

$Cue_{e}-validity$ provides the \textit{centrality} of an entity in a given KBS, by summing all its properties cues. More this value is high, more the entity type maximize the number of its properties with the members it categorises.

Given the notion of $Cue_{e}-validity$, we capture the level of \textit{minimisation of the number of properties shared with other entity types, inside a KBS} with the notion of $cue_{er}-validity$, which we define as:

\vspace{-0.2cm}
\begin{equation}
Cue_{er}(e) = \frac{Cue_{e}(e)}{|prop(e)|} = c \in [0, 1]
\vspace{-0.05cm}
\end{equation}

\begin{figure*}
  \centering
  \includegraphics[width=\linewidth]{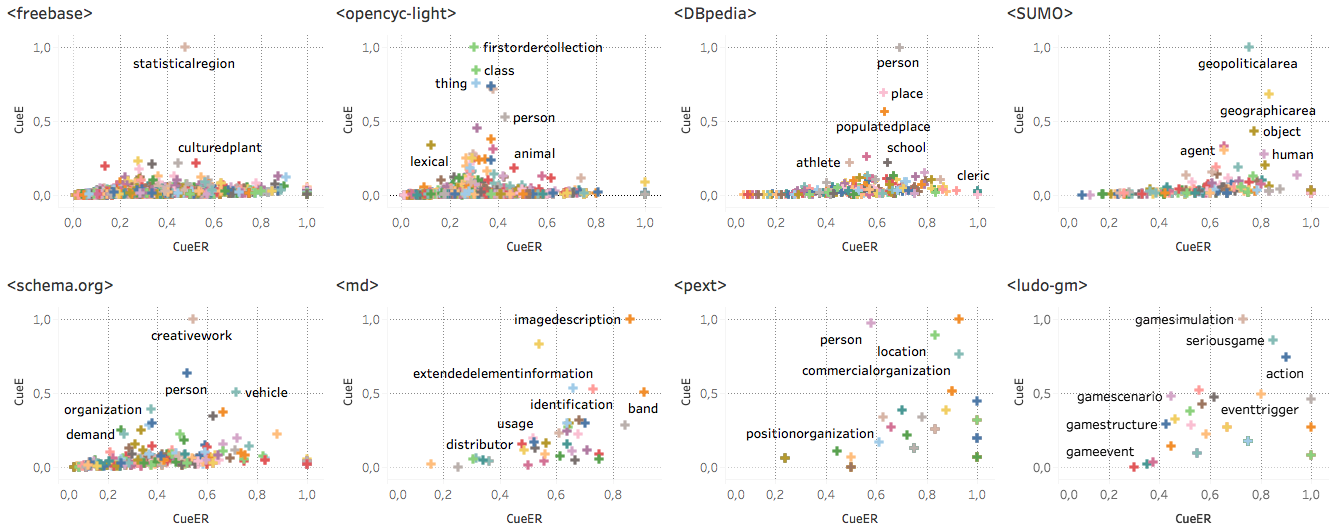}
  \caption{\textit{Entity types categorization relevance for eight example KBSs}}
  \label{02}
  \vspace{-0.2cm}
\end{figure*}

The notions and terminology used for entity types, i.e., the notions of $Cue_{e}$ and $Cue_{er}$, can be straightforwardly generalized to KBSs, generating the following metrics:  

\vspace{-0.2cm}
\begin{equation}
Cue_{k}(K)= \sum_{i=1}^{|E_{K}|} Cue_{e}(e_{i}) = |prop(K)|
\vspace{-0.05cm}
\end{equation}

The $Cue_{k}(K)$ is calculated as a summation of the cues of all the entity types of a given KBS, i.e., $E_{K}$, and returns the number of the properties of the KBS, i.e., $|prop(K)|$.  
Following the formalization of $Cue_{er}$ we capture the level of \textit{minimisation of the number of properties shared across the entity types inside the schema} with the notion of $cue_{kr}-validity$, which we define as:

\vspace{-0.2cm}
\begin{equation}
Cue_{kr}(K)= |prop(K)|/\sum_{i=1}^{|E_{K}|}prop(e_{i}) = c \in [0, 1]
\vspace{-0.05cm}
\end{equation}

Given the definitions of \textit{Cues} for entity types, we capture \textit{the categorization relevance of entity types}, i.e., the entity types that maximise the amount of properties and minimise the number of shared properties, with the notion of $Focus(e)$, which we define as follows: 

\vspace{-0.2cm}
\begin{equation}
Focus(e) = Cue_{e}^{'}(e) + Cue_{er}^{'}(e)
\vspace{-0.05cm}
\end{equation}

where $Cue_{e}^{'}(e)$ is a \textit{log normalization} \cite{bornemann1981log} of $Cue_{e}(e)$ and $Cue_{er}^{'}(e)$ is a \textit{min-max normalization} \cite{jain2011min} of $Cue_{er}(e)$, considering all the $Cue_{er}(e)$ and $Cue_{e}(e)$ values from a given set of KBSs $\left \{ K \right \}$.

Similarly, given the definitions of \textit{Cues} for KBSs, we capture \textit{the categorization relevance of KBSs}, i.e., the KBSs that maximise the number of maximally informative entity types, with the notion of $Focus(K)$, which we define as follows: 

\vspace{-0.2cm}
\begin{equation}
Focus(K) = Cue_{k}^{'}(K) + Cue_{kr}^{'}(K)
\vspace{-0.05cm}
\end{equation}

where $Cue_{k}^{'}(K)$ is a \textit{log normalization} of $Cue_{k}(K)$ and $Cue_{kr}^{'}(K)$ is a \textit{min-max normalization} of $Cue_{kr}(K)$, considering all the $Cue_{k}(K)$ and $Cue_{kr}(K)$ values from a given set of KBSs $\left \{ K \right \}$. Notice that we used log normalization for $Cue_{e}(e)$ and $Cue_{k}(K)$ because of the wide range of $Cue_{e}(e)$ and $Cue_{k}(K)$ values across the KBSs (e.g., we may have a KBS with  $Cue_{e}(e) = 10$ and a KBS with $Cue_{e}(e) = 300$).

\section{Measuring focus}
\label{sec3}
We started from a data set of around 700 KBSs, expressed in the \textit{Terse RDF Triple Language (Turtle)}\footnote{\url{https://www.w3.org/TR/turtle/.}} format. Most of these resources have been taken from the LOV\footnote{\url{https://lov.linkeddata.es/dataset/lov}} catalog. The remaining ones, see for instance \textit{freebase}\footnote{\url{https://developers.google.com/freebase}} and \textit{SUMO}\footnote{\url{http://www.adampease.org/OP/}} have been added to collect more data. 

\begin{table}
\vspace{-0.2cm}
\caption{\textit{KBSs ranking}}
  \label{tab0}
  \centering
  \small
\begin{tabular}{lccc}
\hline
\textbf{KBS}        & \textbf{$Cue_{s}(S)$} & \textbf{$Cue_{sr}(S)$}   & \textbf{$Focus(K)$} \\ \hline
\textit{freebase:}      & \textbf{8981} & \textbf{0,21} & \textbf{1,15}   \\
\textit{cal:}           & \textbf{46}   & \textbf{0,98} & \textbf{0,92}   \\
\textit{bibo:}          & \textbf{71}   & \textbf{0,97} & \textbf{0,92}   \\
\textit{opencyc-l:}       & \textbf{6266} & \textbf{0,26} & \textbf{0,90}   \\
\textit{swpo:}          & \textbf{87}   & \textbf{0,88} & \textbf{0,83}   \\
\textit{cwmo:}          & \textbf{107}  & \textbf{0,85} & \textbf{0,80}   \\
\textit{eli:}           & \textbf{62}   & \textbf{0,84} & \textbf{0,78}   \\
\textit{ncal:}          & \textbf{103}  & \textbf{0,80} & \textbf{0,75}   \\
\textit{mo:}            & \textbf{124}  & \textbf{0,79} & \textbf{0,74}   \\
\textit{akt:}           & \textbf{106}  & \textbf{0,79} & \textbf{0,74} 
\\ \hline
\end{tabular}
\end{table}

For the sake of the analysis, all the data sets have been flattened into a set of sets of triples (one set per entity type), where each triple encodes information about ``entitytype-property'' associations $I_{K}(e)$ (e.g., the triple ``Person-domainOf-friend'' encodes the ``Person-friend'' $I_{K}(e)$ association. Moreover, in order to generate the final output data sets we processed properties labels via NLP pipeline which performs various steps, including, for instance: \textit{i)} split a string every time a capital letter is encountered (e.g., \textit{birthDate} $\rightarrow$ birth and date); \textit{ii)} lower case all characters; \textit{iii)} filter out stop-words (e.g., \textit{hasAuthor} $\rightarrow$ author). This allowed us to run a more accurate analysis. For instance, if ``Person'' and ``Place'' have properties like ``globalLocationNumberInformation'' and ``LocationNumber'', respectively, by processing the labels like we have done is possible to find some overlapping (see ``location'' and ``number'') otherwise no.

\begin{table}
\vspace{-0.2cm}
\caption{\textit{Entity types ranking}}
  \label{tab1}
\centering
\small
\begin{tabular}{ccccc}
\hline
\textbf{KBS}    & \textbf{entity type}                & \textbf{$Cue_{e}(e)$}   & \textbf{$Cue_{er}(e)$} & \textbf{$Focus(e)$} \\ \hline
\textit{DBpedia:}  & \textit{person}               & \textbf{169,02} & \textbf{0,69}  & \textbf{1,42}  \\
\textit{opencyc-l:}  & \textit{firstordercoll.} & \textbf{230,59} & \textbf{0,30}  & \textbf{1,30}  \\
\textit{freebase:} & \textit{statisticalreg.}    & \textbf{161,53} & \textbf{0,48}  & \textbf{1,17}  \\
\textit{opencyc-l:}  & \textit{class}                & \textbf{194,95} & \textbf{0,31}  & \textbf{1,15}  \\
\textit{dicom:}    & \textit{ieimage}              & \textbf{158,90} & \textbf{0,44}  & \textbf{1,13}  \\
\textit{DBpedia:}  & \textit{place}                & \textbf{116,97} & \textbf{0,63}  & \textbf{1,13}  \\
\hline
\end{tabular}
\end{table}

\begin{table}
\vspace{-0.2cm}
\caption{\textit{KBSs ranking for the entity type person}}
  \label{tab2}
\centering
\small
\begin{tabular}{ccccc}
\hline
\textbf{KBS}    & \textbf{entity type}  & \textbf{$Cue_{e}(e)$}   & \textbf{$Cue_{er}(e)$} & \textbf{$Focus(e)$} \\ \hline
\textit{DBpedia:}  & \textit{person} & \textbf{169,02} & \textbf{0,69}  & \textbf{1,42}  \\
\textit{akt:}      & \textit{person} & \textbf{8,00}   & \textbf{1,00}  & \textbf{1,03}  \\
\textit{opencyc-l:}  & \textit{person} & \textbf{122,14} & \textbf{0,43}  & \textbf{0,95}  \\
\textit{vivo:}     & \textit{person} & \textbf{10,60}  & \textbf{0,88}  & \textbf{0,92}  \\
\textit{swpo:}     & \textit{person} & \textbf{3,50}   & \textbf{0,88}  & \textbf{0,88}  \\
\textit{cwmo:}     & \textit{person} & \textbf{5,83}   & \textbf{0,83}  & \textbf{0,85}  \\
 \hline
\end{tabular}
\end{table}

Due to the lack of space, after the above processing we selected a subset of the starting data set, by discharging all the KBSs with less than 30 entity types. An overall view of the final output data set is provided by Figure \ref{01}, where, for each of the 44 KBSs, the number of properties, the number of entity types and the balance are provided. The balance returns the value of a simple distribution of the properties of a KBS across its entity types and it is calculated as $\frac{|prop(K_{i})|}{|E_{K_{i}}|} \ast \frac{1}{|prop(e_{i})|_{max}}$, with $|prop(K_{i})|$ being the cardinality of the set of properties of the KBS, $|E_{K_{i}}|$ being the cardinality of the set of entities of the KBS and $|prop(e_{i})|_{max}$ being the cardinality of the set of properties associated to the entity with the major number of properties in the KBS.

By applying the cue entity metrics, i.e., $Cue_{e}(e)$ and $Cue_{er}(e)$ to the KBSs of the resulting list, we obtained the scores to evaluate the categorization relevance of the entity types for each KBS. Let us take, for instance the values provided by KBSs in (Fig. 2). We randomly selected eight KBSs from the starting set and we listed them according to the number of entity types. The selected KBSs are: \textit{freebase}, \textit{opencyc-light}\footnote{\url{https://pythonhosted.org/ordf/ordf_vocab_opencyc.html}}, \textit{DBpedia}, \textit{SUMO}, \textit{schema.org}, \textit{md}\footnote{\url{http://def.seegrid.csiro.au/isotc211/iso19115/2003/metadata}}, \textit{pext}\footnote{\url{http://www.ontotext.com/proton/protonext.html}} and \textit{ludo-gm}\footnote{\url{http://ns.inria.fr/ludo/v1/docs/gamemodel.html}}. The corresponding scatter plots provide the correlations between (a \textit{min-max normalization} of) $Cue_{e}(e)$ and $Cue_{er}(e)$ for each entity type of each of the selected KBSs. The top-right entity types are the ones with the higher categorization relevance according to our metrics. For instance, in \textit{SUMO} we have entity types like \textit{GeopoliticalArea} and \textit{GeographicalArea} and in \textit{DBpedia} we have \textit{Person} and \textit{Place}.

By applying the $Focus(K)$ over the set of 44 KBSs we obtained the KBS ranking, where the top 11 KBSs are reported in Tab. 1. By applying $Focus(e)$ over the set of 44 KBSs we obtained the entity types ranking, where the top 6 entity types in terms of categorization relevance are reported in Tab. 2. Finally by selecting a given entity type, by applying $Focus(e)$, it is possible to find the best KBS for that entity type. Table 3 provides an example for the entity type \textit{Person}.

\section{Focus evaluation}
\label{sec4}

We organize the evaluation in two parts. In Section \ref{FocusE} we analyse the accuracy of the $Focus(e)$ metric in weighting the categorization relevance of entity types, namely their centrality in the maximization of information. This will be done by applying our metrics and some related SoA ranking algorithms over a set of example KBSs.
Then we compare the results with a reference data set generated by 5 knowledge engineers, to which we provided a set of instructions/ guidelines to rank the entity types, taking inspiration from Rosch's experiment \cite{rosch1999principles}. 

In Section \ref{FocusC}, given the lack of baseline metrics for calculating the overall score of a KBS on similar functions, and the lack of reference gold standards, we analyse the effects that the $Focus(K)$ of a KBS may have on the prediction performance of a relational classification task, thus showing a possible practical application of the given measure. 

\subsection{Evaluating Focus(e)}
\label{FocusE}
\begin{figure}[h]
  \centering
  \includegraphics[width=\linewidth]{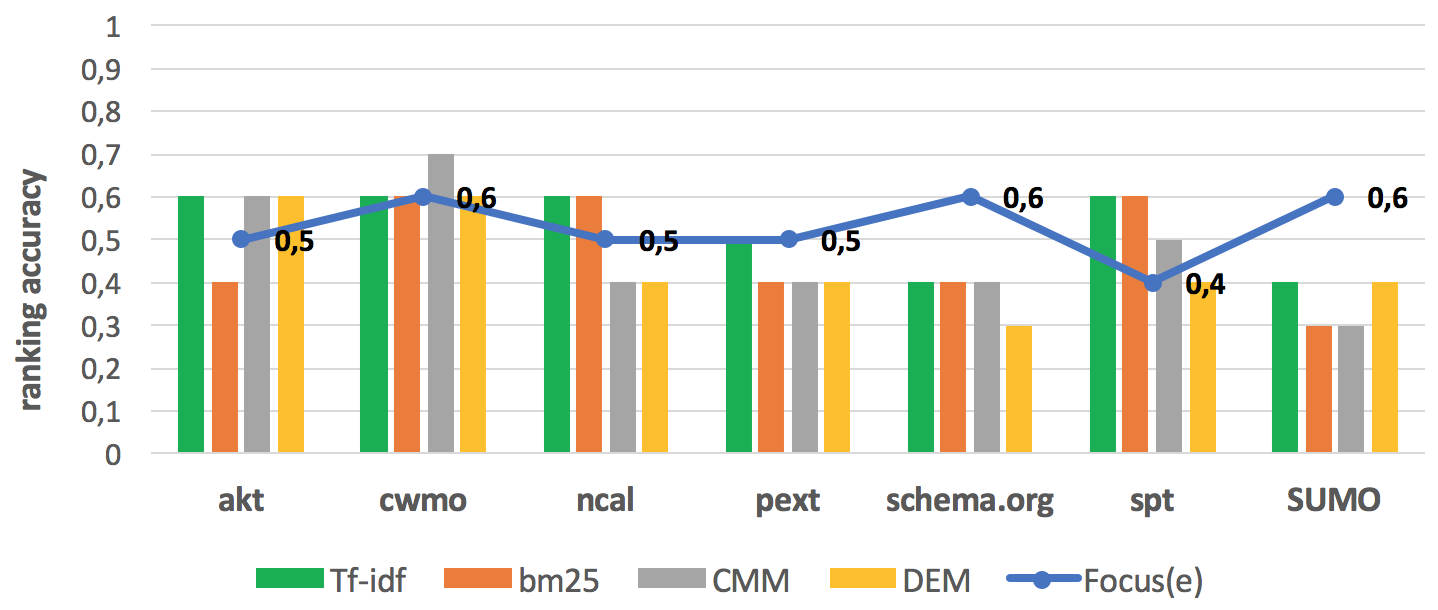}
  \caption{$Focus(e)$ experiment results}
  \label{fig3}
  \vspace{-0.3cm}
\end{figure}

The question here is whether $Focus(e)$ \textit{allows to rank entity types in KBSs according to their categorization relevance}, as described in Section \ref{sec1}. To evaluate our metric we firstly selected a subset of the KBSs discussed in the previous section, namely \textit{akt}\footnote{\url{https://lov.linkeddata.es/dataset/lov/vocabs/akt}}, \textit{cwmo}\footnote{\url{https://gabriel-alex.github.io/cwmo/}}, \textit{ncal}, \textit{pext}, \textit{schema.org}, \textit{spt}\footnote{\url{https://github.com/dbpedia/ontology-tracker/tree/master/ontologies/spitfire-project.eu}} and \textit{SUMO}. We selected these KBSs because they provide very different examples in terms of number of properties, entities, balance and cues. Moreover almost all their entity types labels are human understandable\footnote{A lot of KBSs have entity types labels codified by an ID.}. As second step we selected four SoA ranking algorithms, namely \textit{TF-IDF} \cite{salton1988term}, \textit{BM25} \cite{robertson1995okapi}, \textit{Class Match Measure (CMM)} and \textit{Density Measure (DEM)} \cite{alani2006metrics}. We used the performance of these rankings as baseline, by selecting their scores for the top 10 entity types, for each of the given KBSs, and we compared them with the rankings provided by $Focus(e)$. The relevance of our approach was then measured in terms of accuracy (from 0 to 1) by checking how many entity types of the ranking results are in the entity types ranking lists provided by the knowledge engineers. The output of this experiment is represented by the data in Figure \ref{fig3}.

\vspace{-0.2cm}

Looking at the chart, the line represents the accuracy of the ranking trend provided by our $Focus(e)$. Each bar represents the accuracy of the ranking, for each selected SoA algorithm. All the accuracy results are grouped w.r.t. the reference KBS. 

The first main observation is that all the reference SoA metrics show a very similar trend, with higher accuracy for \textit{akt}, \textit{cwmo}, \textit{ncal} and \textit{spt}, and lower accuracy for \textit{schema.org} and \textit{SUMO}. This is not the case for $Focus(e)$. Our metric, indeed, even if it is not the best for all the KBSs, performs best with huge and very noisy (with lower $Cue_{cr}$, and then lower entity types $Cue_{er}$) KBSs, as it is the case for \textit{schema.org} and \textit{SUMO} (just check the visualization of \textit{SUMO} and \textit{schema.org} as in Figure \ref{02} to observe the phenomenon). This, as we expected, depends on the pivotal role we gave to the minimization of the number of overlapping properties. The $Cue_{er}$ for each entity type provides indeed essential information about the categorization relevance that, giving more importance to the number of properties of an entity type, may not be properly identified. Thus, given small and not-noisy (or ``clean'') KBSs, other approaches, very focused on the number of properties of entity types pay very well (see the good performance of the \textit{TF-IDF} algorithm). Differently, when KBSs present a huge amount of entity types, with low $Cue_{er}$, $Focus{e}$ allows to identify better the categorization relevance. 

The second main observation is that \textit{TF-IDF} and $Focus(e)$ are the best metrics in terms of average performance, namely 0.52 (both \textit{TF-IDF} and $Focus(e)$) mean accuracy vs. 0.47 for \textit{bm25} and \textit{CMM}, and 0.44 for \textit{DEM}. This score being motivated by the fact that \textit{TF-IDF} is almost always the best when the given KBS is small and ``clean'' and $Focus(e)$ compensates the standard performance with small and clean KBSs, with a high performance with huge and noisy KBSs.

\subsection{Evaluating Focus(K)}
\label{FocusC}

The question here is whether Focus(K) helps to predict \textit{the performance of KBSs in their ability to predict their own entity types}. In this experiment we used the same KBSs we selected in the previous experiment to address relational classification, where entity types have an associated label and the task is to predict those labels. Notice that we addressed a specific type of relational classification, namely an \textit{entity type recognition task (ETR)}, as defined in \cite{giunchiglia2020entity}. We set-up the experiment as follows: \textit{i)} we trained a \textit{decision tree} and a \textit{k-NN} \cite{kaminski2018framework,dasarathy1991nearest} model with each KBS converted into FCA format (see details provided in Section \ref{sec3}) and performed standard nested cross-validation (with 50 folds); \textit{ii)} we reported the relative performance of the models in terms of differences in accuracy and compared the performances with the $Focus(K)$ for each of the given KBSs. The results are as in Figure~\ref{focusCex} below.

\vspace{-0.2cm}
\begin{figure}[h]
  \centering
  \includegraphics[width=\linewidth]{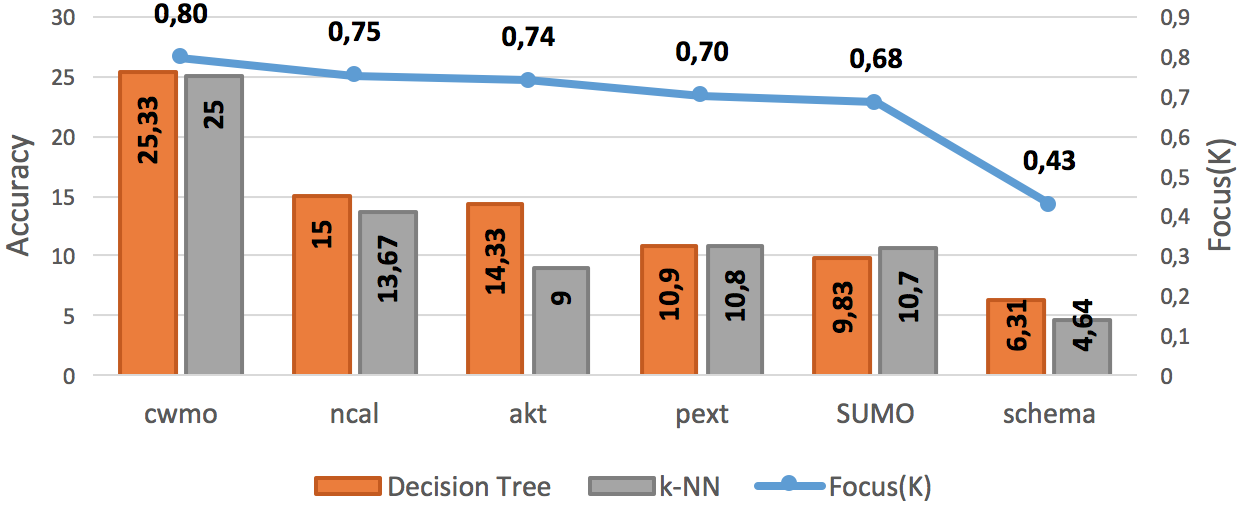}
  \caption{$Focus(K)$ experiment results}
  \label{focusCex}
  \vspace{-0.3cm}
\end{figure}

Looking at the chart, the accuracy is reported as a proportion of correct predictions, within the range of [0\%,100\%] (see charts bars). The $Focus(K)$ is reported by the values of the line. The \textit{cwmo} KBS is the one with the best scores, in terms of accuracy (for both the trained models) and $Focus(K)$. \textit{schema.org} is the worst.

The main observation is that, as expected, the trend in terms of accuracy, considering both the two models, follows the $Focus(K)$ ranking for most of the given KBSs. However, it can be noticed that k-NN, with the \textit{pext} KBS represents an exception, it is indeed worse than \textit{akt} in terms of $Focus(K)$, but performs better with k-NN. Going deep into the analysis, this phenomenon can be explained by the relationship between the number of properties and the number of entity types, more specifically by the balance of the KBS. This value can indeed affect the performance of the model in prediction. The more the balance the more the probability of having entity types with a low focus. This effect being quite evident if we consider two KBSs with very similar $Focus(K)$, but very different balance.

This experiment, while showing how $Focus(K)$ can be a concrete explanation of the categorization relevance of a KBS, suggests the  possibility of a practical application of $Focus(K)$ to evaluate the potential performance of a KBS or a set of KBSs in a relational classification task. The results may be used, e.g., to fine-tune KBSs in an open-world data integration scenario.

\section{Related work}
\label{sec5}
This work shares with the work on {\it ontology} and {\it knowledge graph (KG) schema} (functional) evaluation \cite{mcdaniel2019evaluating,paulheim2017knowledge,brank2005survey,gangemi2005theoretical} the goal of facilitating the reuse of these knowledge structures. This work has been extensive and has exploited a huge amount of methods and techniques including, e.g. \textit{DWRank} \cite{butt2016dwrank} and the \textit{NCBO} \cite{martinez2017ncbo} (the former being a high precision recommender for biomedical ontology, the latter being a ``learning to
rank approach'' based on search queries).

Our work differs from this in two major respects. The first is that we ground our approach and the notion of focus on the notion of
categorization purpose from cognitive psychology. The  theoretical underpinning of our
formalization of the metrics and the experimental setup, is then inspired by the analysis of human behavior in categorization, and in particular by the seminal work by E. Rosch. Our goal is not to redefine terminology already in  use in  the related work,  but  rather  to  propose  a  both  theoretically and practically useful formalization of the central activity of categorization, which can be considered as the baseline of each knowledge engineering task. The second difference, which is actually a consequence of the first, is that, while most of the functional evaluation approaches are related to the intended use of a given KBS, and consider functional dimensions, like task and domain, which are very context dependent, this is not the case with our approach. The notion of focus we adapted, indeed, aims to model a privileged level of categorization, independently from the tasks and the domain of application of the data structure. This in turn allows us to devise a somewhat opposite approach. In fact, the domain of a KBS can be then identified through the focus scores. For instance, the fact that a KBS has a high focus for entity types like \textit{CreativeWork} or \textit{Product}, will help the user to understand what is the real potential of that KBS for a given domain of application.   

As last consideration, it is important to observe how the notion of cue validity has been widely studied in the context of feature engineering. Together with  other similar measures as ``category utility'' or ``mutual information'' and, it has been used to measure the informativeness of a category \cite{peng2005feature}. Our approach differs from the related work in the application of Rosch's notion at the KBS level, rather than on data. Moreover, the introduction of the ``overall'' \textit{Focus} metrics to rank categorization relevance is a novel contribution.

\section{Conclusion}
\label{sec6}
In this paper, we have proposed a formal method to evaluate KBSs according to their focus, namely, what cognitive psychologists call categorization purpose. This in turn has allowed us to describe how this evaluation plays an important role in supporting an accurate level of KBSs understanding and reuse.  
The future work will concentrate on an extension of the proposed metrics, possibly by considering the hierarchical structure of KBSs, an extension of the experimental set-up and an implementation of the metrics for supporting the search engine of a large number of existing high-quality KBSs.

\newpage

\bibliographystyle{named}
\bibliography{ijcai20}

\end{document}